\documentclass[10pt,twocolumn,letterpaper]{article}

\usepackage[pagenumbers]{cvpr} 

\usepackage{lipsum}
\usepackage{array}
\usepackage{times}
\usepackage{epsfig}
\usepackage{graphicx}
\usepackage{float}
\usepackage{wrapfig}
\usepackage{amsmath,amssymb,amsthm}
\usepackage{algorithm,algorithmicx,algpseudocode}
\usepackage{bm,xspace}
\usepackage{comment}
\usepackage{multirow}
\usepackage{balance}
\usepackage{url}
\usepackage{booktabs}
\usepackage{etoolbox,siunitx}
\usepackage{calc}
\usepackage{pifont,hologo}
\usepackage{color}
\usepackage{adjustbox}
\usepackage[normalem]{ulem}  
\usepackage[dvipsnames]{xcolor}
\usepackage[pagebackref,breaklinks,colorlinks,linkcolor=red,citecolor=blue]{hyperref}
\usepackage{natbib}
\usepackage{arydshln}
\usepackage{cuted}
\PassOptionsToPackage{square,numbers,comma,sort}{natbib}

\definecolor{blue}{HTML}{0055cc}
\definecolor{red}{HTML}{cc1100}
\definecolor{orange}{HTML}{cc7700}
\definecolor{gray}{HTML}{efefef}
\definecolor{darkgreen}{rgb}{0.13, 0.55, 0.13}
\definecolor{darkgray}{HTML}{757575}

\newcommand{\figref}[1]{Figure~\ref{#1}}
\newcommand{\tabref}[1]{Table~\ref{#1}}
\newcommand{\secref}[1]{Section~\ref{#1}}
\renewcommand{\eqref}[1]{Eq.~\ref{#1}}

\newcolumntype{P}[1]{>{\centering\arraybackslash}p{#1}}
\newcolumntype{M}[1]{>{\centering\arraybackslash}m{#1}}
\newcolumntype{x}[1]{>{\centering\arraybackslash}p{#1}}
\newcolumntype{y}[1]{>{\raggedright\arraybackslash}p{#1}}
\newcolumntype{z}[1]{>{\raggedleft\arraybackslash}p{#1}}
\newcommand{\tablestyle}[2]{\setlength{\tabcolsep}{#1}\renewcommand{\arraystretch}{#2}\centering\footnotesize}

\DeclareMathSymbol{@}{\mathord}{letters}{"3B}

\makeatletter
\DeclareRobustCommand\onedot{\futurelet\@let@token\@onedot}
\def\@onedot{\ifx\@let@token.\else.\null\fi\xspace}

\newcommand*{\Rom}[1]{\expandafter\@slowromancap\romannumeral #1@}
\newcommand*{\rom}[1]{\expandafter\romannumeral #1}

\def\1{\bm{1}}

\def\rvx{{\mathbf{x}}}

\DeclareMathAlphabet{\mathsfit}{\encodingdefault}{\sfdefault}{m}{sl}
\SetMathAlphabet{\mathsfit}{bold}{\encodingdefault}{\sfdefault}{bx}{n}

\usepackage[accsupp]{axessibility} 

\def\paperID{7971} 
\def\confName{CVPR}
\def\confYear{2024}

\title{DreamComposer: Controllable 3D Object Generation via Multi-View Conditions}

\author{Yunhan Yang$^{1\ast}$
\qquad
Yukun Huang$^{1\ast}$
\qquad
Xiaoyang Wu$^{1}$
\qquad
Yuan-Chen Guo$^{3,4}$
\\
Song-Hai Zhang$^{4}$
\qquad
Hengshuang Zhao$^{1}$
\qquad
Tong He$^{2}$
\qquad
Xihui Liu$^{1\dag}$
\vspace{0.2cm}
\\
{\normalsize $^{1}$ The University of Hong Kong \quad $^{2}$ Shanghai Artificial Intelligence Lab}
{\normalsize \quad $^{3}$ VAST \quad $^{4}$ Tsinghua University}\\
{\normalsize $^{\ast}$ Equal Contribution \qquad Project Page: {\href{https://yhyang-myron.github.io/DreamComposer/}{https://yhyang-myron.github.io/DreamComposer/}}}
}

\begin{document}
\maketitle

\let\thefootnote\relax\footnotetext{$^{\dag}$ Corresponding author.}

\begin{strip}
    \centering
    \vspace{-4em}
    \includegraphics[width=\textwidth]{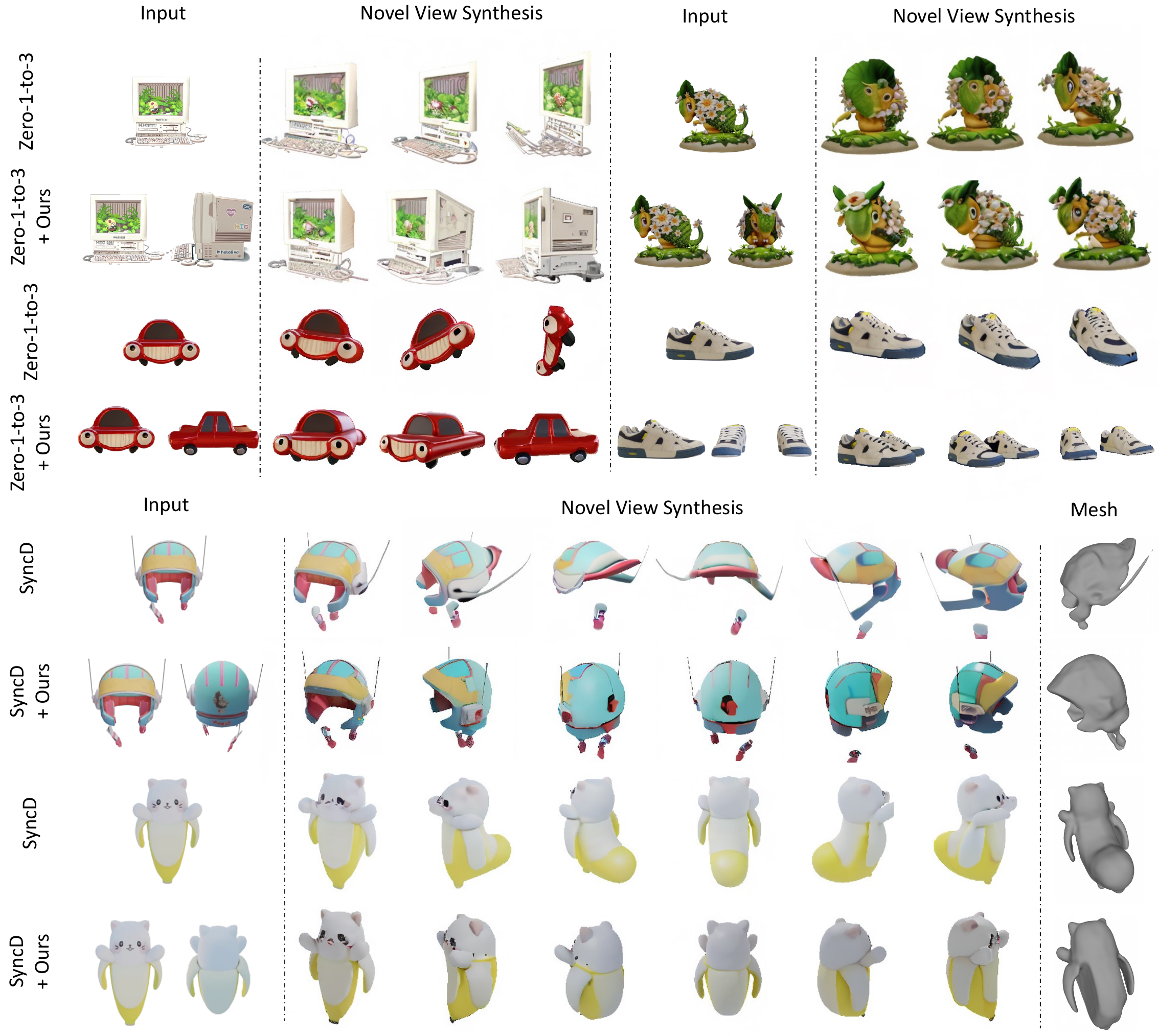}
    \captionof{figure}{\textbf{DreamComposer} is able to generate controllable novel views and 3D objects via injecting multi-view conditions. We incorporate the method into the pipelines of Zero-1-to-3~\cite{liu2023zero1to3} and SyncDreamer (SyncD)~\cite{liu2023syncdreamer} to enhance the control ability of those models.}
    \label{fig:teaser}
\end{strip}

\begin{abstract}
  Utilizing pre-trained 2D large-scale generative models, recent works are capable of generating high-quality novel views from a single in-the-wild image. However, due to the lack of information from multiple views, these works encounter difficulties in generating controllable novel views. In this paper, we present DreamComposer, a flexible and scalable framework that can enhance existing view-aware diffusion models by injecting multi-view conditions. Specifically, DreamComposer first uses a view-aware 3D lifting module to obtain 3D representations of an object from multiple views. Then, it renders the latent features of the target view from 3D representations with the multi-view feature fusion module. Finally the target view features extracted from multi-view inputs are injected into a pre-trained diffusion model. Experiments show that DreamComposer is compatible with state-of-the-art diffusion models for zero-shot novel view synthesis, further enhancing them to generate high-fidelity novel view images with multi-view conditions, ready for controllable 3D object reconstruction and various other applications.

\vspace{-10pt}

\end{abstract}

\section{Introduction}
\label{sec:intro}

3D object generation is an emerging research topic in computer vision and graphics, serving a wide range of applications such as augmented reality (AR), virtual reality (VR), film production, and game industry. With 3D object generation models, users and designers can easily create the desired 3D assets with text or image prompts, without requiring considerable human endeavors by human experts.

Recently, diffusion models~\cite{ho2020denoising, rombach2021highresolution} achieve remarkable success in generating 2D images from texts, which inspires the exploration of 3D object generation using 2D diffusion priors~\cite{poole2022dreamfusion, wang2023score, huang2023dreamtime, wang2023prolificdreamer, lin2023magic3d, chen2023fantasia3d}. Although great 3D generation results have been achieved~\cite{wang2023prolificdreamer}, 2D diffusion models lack view control and struggle to provide view-consistent supervision, resulting in various quality issues of 3D generation such as multiple faces and blurry details. To alleviate this problem, Zero-1-to-3~\cite{liu2023zero1to3} empowers 2D diffusion models with viewpoint conditioning, enabling zero-shot novel view synthesis (NVS) conditioned on a single-view image and image-to-3D object generation. Considering the inconsistent output of Zero-1-to-3, a series of subsequent works~\cite{liu2023syncdreamer, shi2023zero123plus, ye2023consistent1to3, weng2023consistent123, long2023wonder3d} are proposed to improve the 3D consistency of the generated multi-view images.
However, limited by the incomplete information of single-view input, these methods inevitably encounter unpredictable and implausible shapes and textures when predicting novel views. For example, as shown on the right side of the third row of~\figref{fig:teaser}, the actual number of shoes cannot be determined if only given a side view of the shoes. In other words, novel view synthesis and 3D object generation are not fully controllable with only single-view image conditions.

To address this problem, our core idea is to introduce flexible multi-view image conditioning to diffusion models, enabling more controllable novel view synthesis and 3D object reconstruction. For example, based on the front view, back view, and side view of an object drawn by designers, the model will generate images of other viewpoints that are consistent with the multiple input images. It also allows interactive 3D generation where users can provide conditioning images from new viewpoints if the generated 3D objects do not follow the user intention. However, such an attempt is challenging for two reasons. Firstly, it is non-trivial to integrate arbitrary numbers of input views into consistent 3D representations that can guide the generation of the target view image. 
Secondly, it is challenging to design a flexible framework that is compatible with and can be plugged into existing models such as Zero-1-to-3~\cite{liu2023zero1to3} and SyncDreamer~\cite{liu2023syncdreamer} to empower multi-view conditioning for various models.

To this end, we propose DreamComposer, a scalable and flexible framework that can extend existing view-conditioned models to adapt to an arbitrary number of multi-view input images. DreamComposer comprises three stages: target-aware 3D lifting, multi-view feature fusion, and target-view feature injection. 
(i) Target-Aware 3D Lifting encodes multi-view images into latent space and then lifts the latent features to 3D tri-planes~\cite{chan2022efficient}. The tri-plance representation with latent features is compact and efficient, and the target-view-aware 3D lifting design allows the network to focus more on building 3D features related to the target view. (ii) Multi-View feature fusion renders and fuses the 3D features from different views to target-view 2D features with a novel composited volume rendering approach. (iii) Target-View Feature Injection injects the latent features from the previous stage into the diffusion models with a ControlNet-like structure. The injection module takes the relative angle as condition, allowing for adaptive gating of multi-view conditions. DreamComposer can be flexibly plugged into existing models, such as Zero-1-to-3~\cite{liu2023zero1to3} and SyncDreamer~\cite{liu2023syncdreamer}, and endow them with the ability to handle multi-view input images, as shown in~\figref{fig:teaser}.

In summary, we propose DreamComposer, a scalable and flexible framework to empower diffusion models for zero-shot novel view synthesis with multi-view conditioning. The scalability and flexibility of DreamComposer are empowered by our novel design of the target-aware 3D lifting, multi-view feature fusion, and target-view feature injection modules. Extensive experiments show that DreamComposer is compatible with recent state-of-the-art methods, endowing high-fidelity novel view synthesis, controllable 3D object reconstruction, and various other applications such as controllable 3D object editing and 3D character modeling with the ability to take multi-view inputs.

\section{Related Work}
\label{sec:related}

\noindent\textbf{Zero-shot Novel View Synthesis.} Previous works~\cite{mildenhall2021nerf,kulhanek2022viewformer,gu2023nerfdiff} on novel view synthesis are generally trained on datasets with limited scenes or categories and cannot generalize to in-the-wild image inputs.
Recently, diffusion models~\cite{rombach2021highresolution,saharia2022imagen} trained on large-scale Internet data have demonstrated powerful open-domain text-to-image generation capabilities. This success inspired the community to implement zero-shot novel view synthesis by fine-tuning these pre-trained diffusion models. Zero-1-to-3~\cite{liu2023zero1to3} fine-tuned the Stable Diffusion model~\cite{rombach2021highresolution} on the large 3D dataset Objaverse~\cite{deitke2022objaverse}, achieving viewpoint-conditioned image synthesis of an object from a single in-the-wild image.
Based on Zero-1-to-3, several subsequent works~\cite{liu2023syncdreamer, shi2023zero123plus, ye2023consistent1to3, weng2023consistent123, long2023wonder3d, liu2023one2345} aim to produce multi-view consistent images from a single input image to create high-quality 3D objects. However, limited by the ambiguous information of single input image, these models might produce uncontrollable results when rendering novel views.

\noindent\textbf{Diffusion Models for Novel View Synthesis.} In addition to fine-tuning directly on the pre-trained text-image diffusion models,
Some recent works~\cite{zou2023sparse3d, zhou2023sparsefusion, kulhanek2022viewformer, watson2022novel, chan2023generative, gu2023nerfdiff} also attempt to combine diffusion models with 3D priors for novel view synthesis. GeNVS~\cite{chan2023generative} integrates geometry priors in the form of a 3D feature volume into the 2D diffusion backbone, producing high-quality, multi-view-consistent renderings on varied datasets. NerfDiff~\cite{gu2023nerfdiff} distills the knowledge of a 3D-aware conditional diffusion model into NeRF at test-time, avoiding blurry renderings caused by severe occlusion. While remarkable outcomes have been obtained for particular object categories from ShapeNet~\cite{chang2015shapenet} or Co3D~\cite{reizenstein2021common}, the challenge of designing a generalizable 3D-aware diffusion model for novel view synthesis from any in-the-wild inputs remains unresolved.

\noindent\textbf{3D Object Generation.} Due to the limited size of existing 3D datasets, it remains challenging to train generative 3D diffusion models~\cite{jun2023shape, nichol2022pointe, wang2022rodin, müller2023diffrf} using 3D data. With pre-trained text-to-image diffusion models and score distillation sampling~\cite{poole2022dreamfusion}, DreamFusion-like methods~\cite{poole2022dreamfusion, wang2023score, wang2023prolificdreamer, lin2023magic3d, chen2023fantasia3d, metzer2022latentnerf, hong2023debiasing, zhu2023hifa, huang2023dreamtime} have achieved remarkable text-to-3D object generation by distilling 2D image priors into 3D representations. Some methods~\cite{melas2023realfusion, tang2023make, xu2023neurallift, qian2023magic123, tang2023makeit3d, xu2023neurallift360} utilize similar distillation approaches to execute image-to-3D tasks. Since these works, which rely on an optimization strategy, have not previously encountered real 3D datasets, they face the Janus (multi-face) problem, making it challenging to generate high-quality 3D object shapes.

\begin{figure*}
\centering
\includegraphics[width=\linewidth]{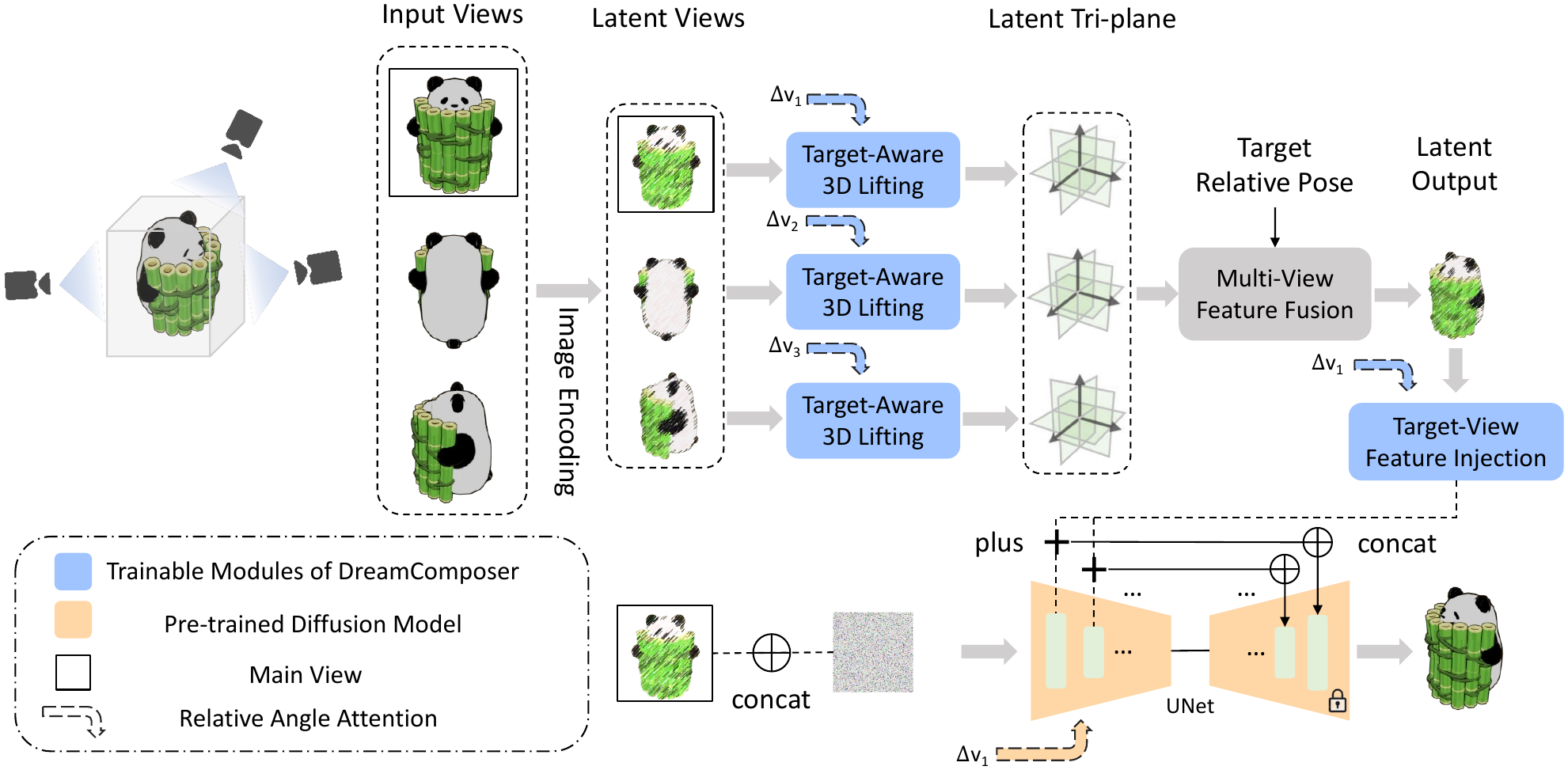}
\caption{An overview pipeline of \textbf{DreamComposer}. Given multiple input images from different views, DreamComposer extracts their 2D latent features and uses a 3D lifting module to produce tri-plane 3D representations. Then, the multi-view condition rendered from 3D representations is injected into the pre-trained diffusion model to provide target-view auxiliary information.}
\vspace{-1em}
\label{fig:pipeline}
\end{figure*}

\section{Method}
\label{sec:method}

DreamComposer aims to empower existing diffusion models for zero-shot novel view synthesis~\cite{liu2023zero1to3,liu2023syncdreamer,ye2023consistent1to3} with multi-view conditions. It consists of three components: (i) \textit{Target-Aware 3D Lifting} extracts 2D features from multi-view inputs and transforms them into 3D representations (Sec.~\ref{sec:3d_lifting}); (ii) \textit{Multi-View Feature Fusion} renders and fuses the 3D features from different views to target-view 2D features with a novel composited volume rendering approach (Sec.~\ref{sec:view_fusion}); (iii) \textit{Target-View Feature Injection} injects the target-view features extracted from multi-view inputs into the diffusion models for multi-view controllable novel view synthesis (Sec.~\ref{sec:injection}). All components are optimized in an Adapter~\cite{houlsby2019parameter,zhang2023adding} fashion (Sec.~\ref{sec:multi_view_training}). An overview pipeline of DreamComposer is demonstrated in \figref{fig:pipeline}.

\noindent\textbf{Formulation.}
Given a main view $\rvx_1 \in \mathbb{R}^{H \times W \times 3}$ and several additional views $\rvx_2, \rvx_3,...,\rvx_n$ of an object, our target is to synthesize the novel view $\hat{\rvx}$ with the relative angle $\Delta \gamma$ to the main view. With the relative angle $\Delta \gamma$, we can calculate the relative camera rotation $R \in \mathbb{R}^{3\times 3}$ and translation $T \in \mathbb{R}^3$. In general, we aim to learn a model $\mathcal{M}$ that can synthesize a novel view $\hat{\rvx}_{R,T}$ from a main view $\rvx_1$ and multiple conditional views $\rvx_2, \rvx_3,...,\rvx_n$:
\begin{align}
    \hat{\rvx}_{R,T} = \mathcal{M}(\rvx_1, \rvx_2, \rvx_3,...,\rvx_n, R, T).
\end{align}

\subsection{Target-Aware 3D Lifting}\label{sec:3d_lifting}
Existing diffusion models~\cite{liu2023zero1to3,liu2023syncdreamer} for zero-shot novel view synthesis are specialized for single-view input and therefore cannot handle an undefined number of multi-view inputs. For a scalable solution, we propose to lift 2D features from different views into 3D representations, ready for view-conditional control.

\noindent\textbf{2D-to-3D Feature Lifting.} Given an input image $\rvx_i \in \mathbb{R}^{H \times W \times 3}$ from the camera view $i$, we first utilize the image encoder of Stable Diffusion~\cite{rombach2021highresolution} to encode it into latent feature $f_i \in \mathbb{R}^{H' \times W' \times 4}$, where $H' \times W'$ is down-sampled image size. Then, we introduce a 3D lifting module with a convolutional encoder structure with self-attention and cross-attention layers. The 3D lifting module lifts the 2D latent feature $f_i$ into a 3D representation $F_i \in \mathbb{R}^{H' \times W' \times 32 \times 3}$ conditioned on the relative angle $\Delta \gamma$. We adopt the tri-plane~\cite{chan2022efficient} feature $F_i = \{F_i^{xy}, F_i^{xz}, F_i^{yz}\}$ as the 3D representation as it is \textit{compact and efficient enough to alleviate the high training cost caused by multi-view inputs}. Note that the 2D-to-3D feature lifting is performed in latent space, which \textit{significantly reduces the computational cost}.

The network structure of the 3D lifting module includes self-attention layers, cross-attention layers, and convolutional layers. Here we design a view conditioning mechanism based on cross-attention, enabling adaptive 3D lifting. Specifically, we take the angle difference between the input view and the target view as a condition and inject it into the 3D lifting module through the cross-attention layers. This mechanism allows 3D lifting to \textit{focus more on building 3D features related to the target view, rather than trying to construct a complete 3D representation.}

\noindent\textbf{Multi-View Cases.} Given multiple input images from $n$ different views, i.e. $\rvx_1, \rvx_2 ,..., \rvx_n$, we can obtain their tri-plane features $\{F_1, F_2 ,..., F_n\}$ via 2D image encoding and 2D-to-3D feature lifting. These tri-plane features are ready for providing target-view auxiliary information in subsequent multi-view conditioning.

\subsection{Multi-View Feature Fusion}\label{sec:view_fusion}

After obtaining the 3D features $\{F_1, F_2 ,..., F_n\}$ of input images from $n$ different views, target-view latent feature $f_t$ can be extracted from these 3D features as the condition for the diffusion model.

To render the target-view latent feature $f_t$, 3D features from different views need to be fused. However, this is tricky because these 3D features are lifted in different camera spaces and are not aligned. To deal with it, we use a composited volume rendering approach: (1) sampling ray points from the target view; (2) projecting these points onto different input-view camera spaces; (3) indexing and aggregating 3D point features from different views; (4) integrating point features along the target-view rays to render the desired latent feature $f_t$.

In particular, we adopt a weighting strategy to adaptively aggregate 3D point features from different inputs, considering that different input views contribute differently to the target view. Given $n$ input views, the azimuth differences between them and the target view are denoted as $\Delta \theta_1, \Delta \theta_2, ..., \Delta \theta_n$. Then, the weight of input view $i$ can be formulated as:
\vspace{-5pt}
\begin{align}
\lambda_i = \frac{\cos{\Delta \theta_i} + 1}{2},
\end{align}
and the weighted 3D point feature aggregation across different views is formulated as:
\begin{align}
f_p^t = \sum_{i=1}^n \bar{\lambda}_i \cdot f_p^i,
\end{align}
where $f_p^t$ and $f_p^i$ denote feature embeddings of 3D point $p$ from target view and input view $i$, respectively; while $\bar{\lambda}_i$ is the nomalized weight of input view $i$ calculated by $\lambda_i / \sum_{i=1}^n \lambda_i$.
Finally, all sampled 3D point's features $f_p^t$ are integrated along the target-view rays using the volume rendering equation~\cite{mildenhall2021nerf} and yield $f_t$.

\subsection{Target-View Feature Injection}\label{sec:injection} 
Latent feature $f_t$ contains rich target-view information extracted from multi-view inputs. We inject $f_t$ into the diffusion model's UNet to provide multi-view conditions. To achieve this, we follow ControlNet~\cite{zhang2023adding} structure for target-view feature injection. Specifically, we clone the network blocks of the diffusion model's UNet to trainable copies. These copies, serving as target-view feature injection modules, take the latent feature $f_t$ as conditional input and predict residuals added to the intermediate outputs of UNet.

Most details are consistent with ControlNet~\cite{zhang2023adding}, except that the input layer needs to be modified to match the size of latent input $f_t$.
Besides, we also take the angle difference between the main
view and the target view as a condition, and inject it into the multi-view injection module through the cross-attention layers.
This design enables adaptive gating of multi-view conditions: auxiliary information from multiple views is less important when the main view and the target view are close.

\subsection{Training and Inference}\label{sec:multi_view_training}

In Sec.~\ref{sec:3d_lifting}, Sec.~\ref{sec:view_fusion}, and Sec.~\ref{sec:injection}, we respectively introduce the target-aware 3D lifting, multi-view feature fusion, and target-view feature injection modules, empowering the pre-trained diffusion model with multi-view inputs. Among those modules the target-aware 3D lifting and target-view injection modules are trainable. To train these additional modules, we always sample three views of objects in each iteration, including a front view, a back view, and a random view. This sampling strategy improves training efficiency while encouraging generalization to arbitrary view inputs. Given multi-view input images, we further propose a two-stage training paradigm.

In the first stage, we pre-train the target-aware 3D lifting module on the proxy task of sparse view reconstruction. Given several input views of an object, the 3D lifting module is encouraged to predict novel views correctly, with a mean square error (MSE) loss in latent space as objective.

In the second stage, a pre-trained diffusion model such as Zero-1-to-3~\cite{liu2023zero1to3} is introduced as the frozen backbone. To enhance it with multi-view conditioning, our target-aware 3D lifting, multi-view feature fusion, and target-view feature injection are integrated and optimized jointly. We use diffusion loss and MSE loss as in the first stage for training.

In the inference stage, the trained model is flexible and can take one or more images from different views as inputs, enabling zero-shot novel view synthesis under multi-view conditions. It also benefits downstream 3D reconstruction and generation tasks with scalability and controllability.

\begin{figure}
    \centering
    \includegraphics[width=1.0\linewidth]{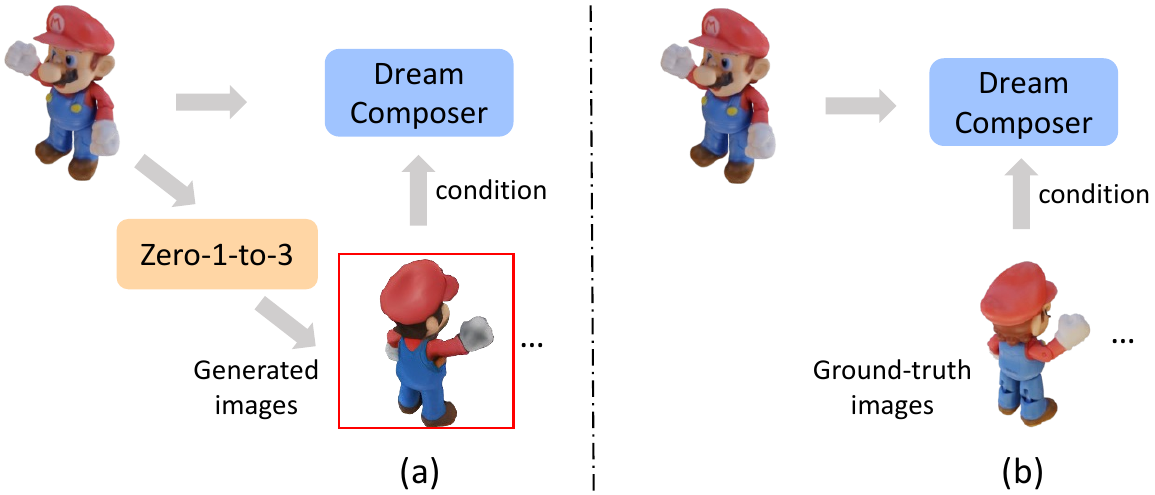}
    \caption{\textbf{Different numbers of ground-truth inputs.} Our model is capable of handling a variety of ground-truth input quantities.}
    \vspace{-1em}
    \label{fig:input-num-vis}
\end{figure}

\begin{figure}
    \centering
    \includegraphics[width=0.9\linewidth]{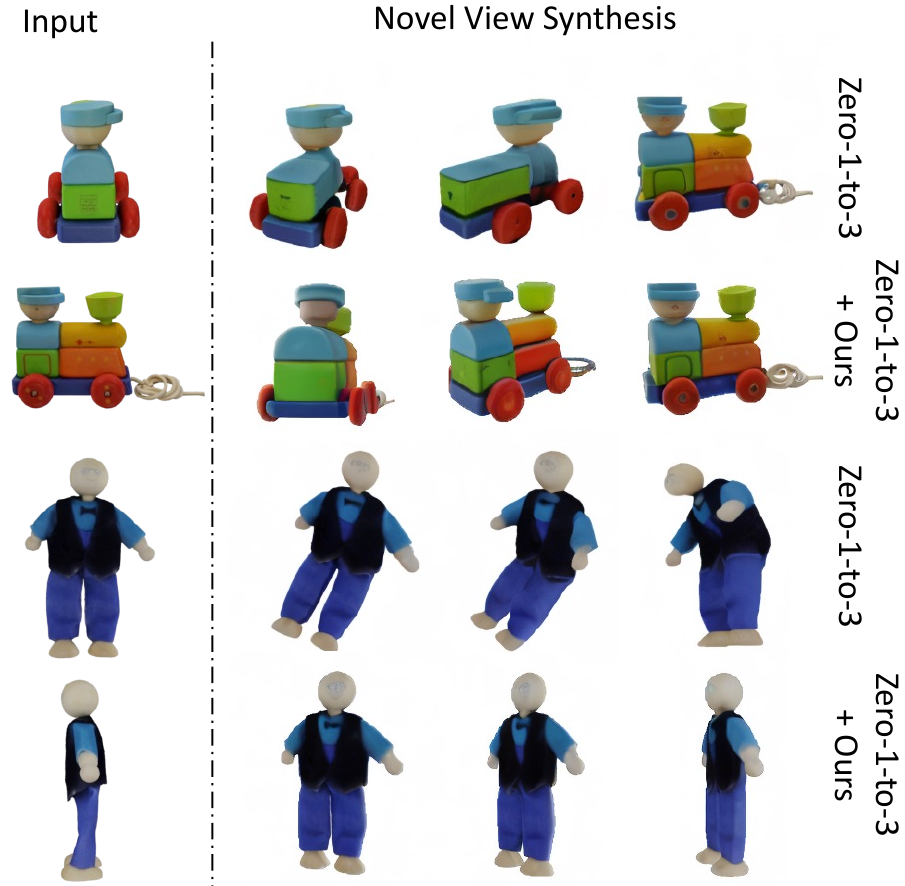}
    \caption{Qualitative comparisons with Zero-1-to-3~\cite{liu2023zero1to3} in controllable novel view synthesis. DC-Zero-1-to-3 effectively generates more controllable images from novel viewpoints by utilizing conditions from multi-view images.}
    \vspace{-1em}
    \label{fig:zero123_nvs}
\end{figure}

\begin{table}
    \begin{minipage}{0.48\textwidth}
    \subcaption{Elevation Degree - 0}\label{subtab:ed0}
    \tablestyle{12pt}{1.08}
    \begin{tabular}{c|ccc}
        Methods & PSNR $\uparrow$ & SSIM $\uparrow$ & LPIPS $\downarrow$ \\
        \toprule
        Zero-1-to-3~\cite{liu2023zero1to3} & 20.82 & 0.840 & 0.139 \\
        Zero-1-to-3+Ours & \textbf{25.25} & \textbf{0.888} & \textbf{0.088} \\
    \end{tabular}
    \end{minipage} \\
    \begin{minipage}{0.48\textwidth}
    \subcaption{Elevation Degree - 15}\label{subtab:ed15}
    \tablestyle{12pt}{1.08}
    \begin{tabular}{c|ccc}
        Methods & PSNR $\uparrow$ & SSIM $\uparrow$ & LPIPS $\downarrow$ \\
        \toprule
        Zero-1-to-3 & 21.38 & 0.837 & 0.131 \\
        Zero-1-to-3+Ours & \textbf{25.85} & \textbf{0.891} & \textbf{0.083} \\
    \end{tabular}
    \end{minipage} \\
    \begin{minipage}{0.48\textwidth}
    \subcaption{Elevation Degree - 30}\label{subtab:ed30}
    \tablestyle{12pt}{1.08}
    \begin{tabular}{c|ccc}
        Methods & PSNR $\uparrow$ & SSIM $\uparrow$ & LPIPS $\downarrow$ \\
        \toprule
        Zero-1-to-3 & 21.66 & 0.837 & 0.128 \\
        Zero-1-to-3+Ours & \textbf{25.63} & \textbf{0.885} & \textbf{0.086} \\
    \end{tabular}
    \end{minipage} \\
    \caption{Quantitative analysis of novel view synthesis using the GSO dataset is presented, employing four distinct angles as inputs. The closest image to the desired viewpoint serves as the input for Zero-1-to-3 and the primary view for DC-Zero-1-to-3, with the remaining three images acting as auxiliary views for DC-Zero-1-to-3. Additionally, we compute results for input elevation angles set at 0, 15, and 30 degrees, respectively.}
    \vspace{-.5em}
    \label{tab:zero123_nvs}
\end{table}

\begin{figure*}
    \centering
    \includegraphics[width=1.0\linewidth]{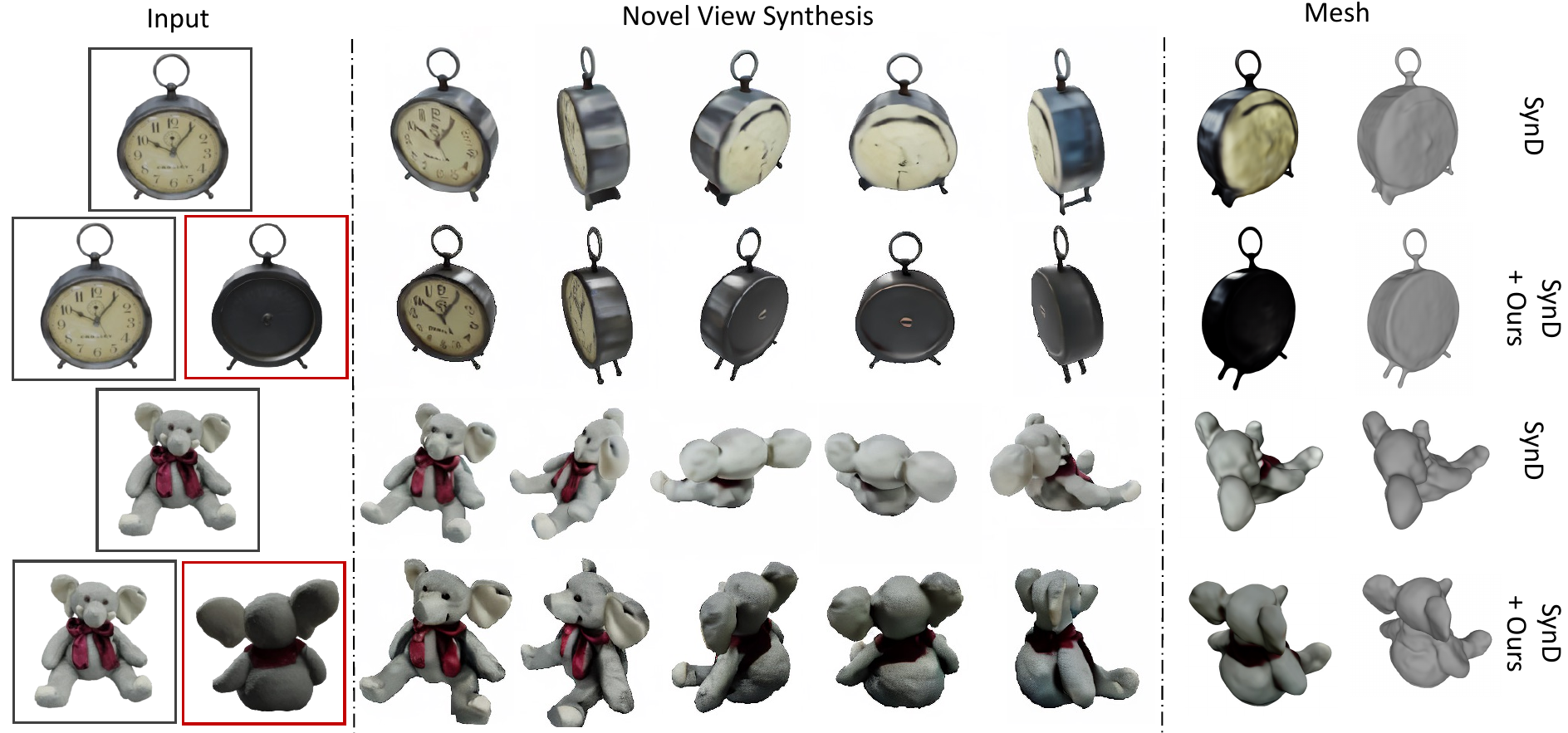}
    \caption{Qualitative comparison with SyncDreamer (SyncD)~\cite{liu2023syncdreamer} in controllable novel view synthesis and 3D reconstruction. The image in $\square$ is the main input, and the other image in \textcolor{red}{$\square$} is the conditional input generated from Zero-1-to-3~\cite{liu2023zero1to3}. With more information in multi-view images, DC-SyncDreamer is able to generate more accurate back textures and more controllable 3D shapes.}
    \vspace{-1em}
    \label{fig:sync_nvs}
\end{figure*}

\begin{table}
    \centering
    \tablestyle{12pt}{1.08}
    \begin{tabular}{c|ccc}
       Method  & PSNR$\uparrow$ & SSIM$\uparrow$ & LPIPS$\downarrow$ \\
       \toprule
       Realfusion~\cite{melas2023realfusion}    
       & 15.26 & 0.722 & 0.283   \\
       Zero-1-to-3~\cite{liu2023zero1to3}    
       & 18.93 & 0.779 & 0.166   \\
       SyncDreamer~\cite{liu2023syncdreamer}    
       & 20.05 & 0.798 & 0.146   \\
       SyncDreamer+Ours 
       & \textbf{20.52} & \textbf{0.828} & \textbf{0.141}   \\
    \end{tabular}
    \caption{Quantitative comparisons of novel view synthesis on GSO dataset. We employ images generated from diffusion models as our additional condition-view.}
    \vspace{-1em}
    \label{tab:sync_nvs}
\end{table}

\begin{figure}
    \centering
    \includegraphics[width=1.0\linewidth]{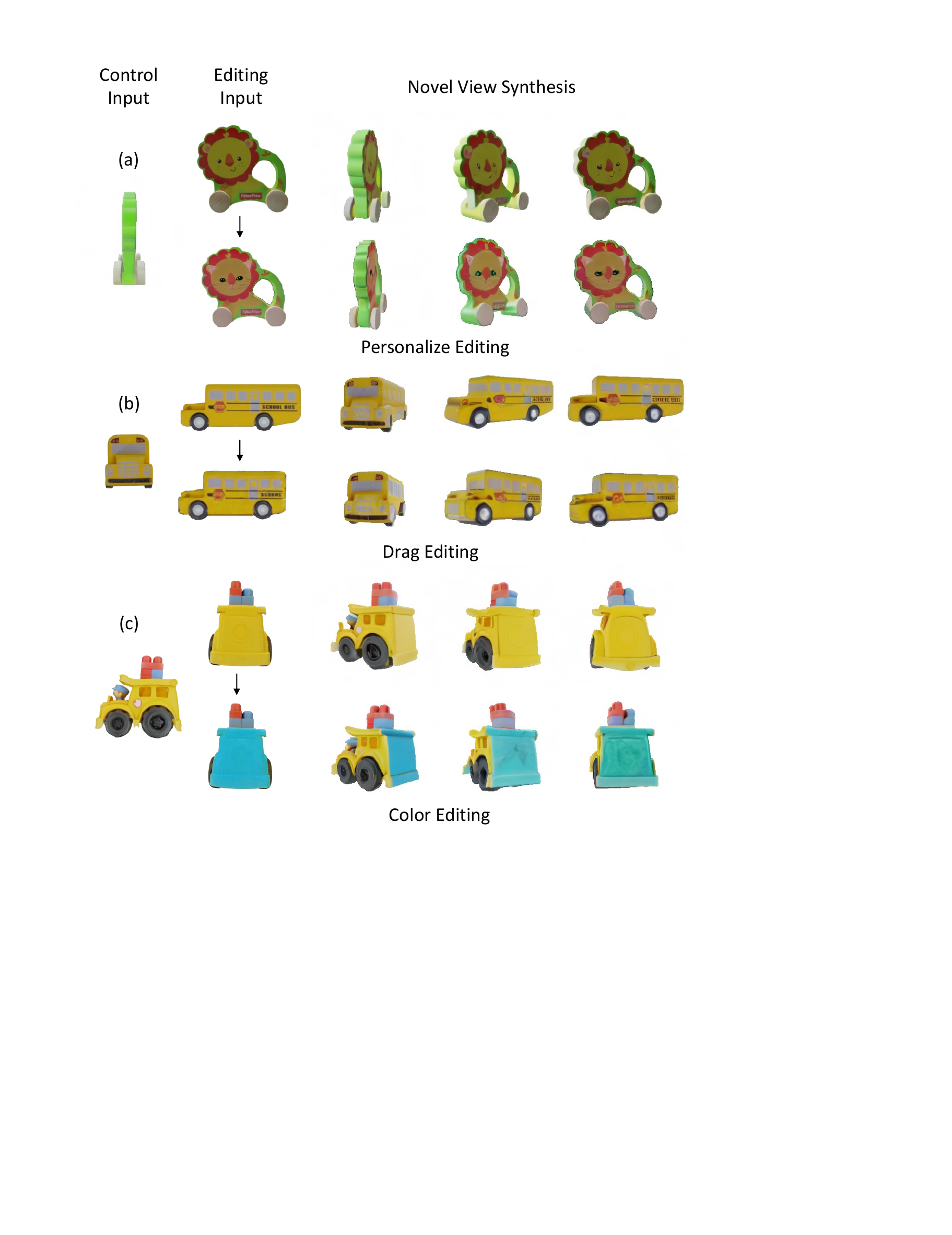}
    \caption{\textbf{Controllable Editing.} We present personalize editing with InstructPix2Pix~\cite{brooks2023instructpix2pix} in (a), drag editing with DragGAN~\cite{pan2023drag}, DragDiffusion~\cite{shi2023dragdiffusion} in (b), and color editing in (c).}
    \vspace{-1em}
    \label{fig:editing}
\end{figure}

\begin{figure*}
    \centering
    \includegraphics[width=1.0\linewidth]{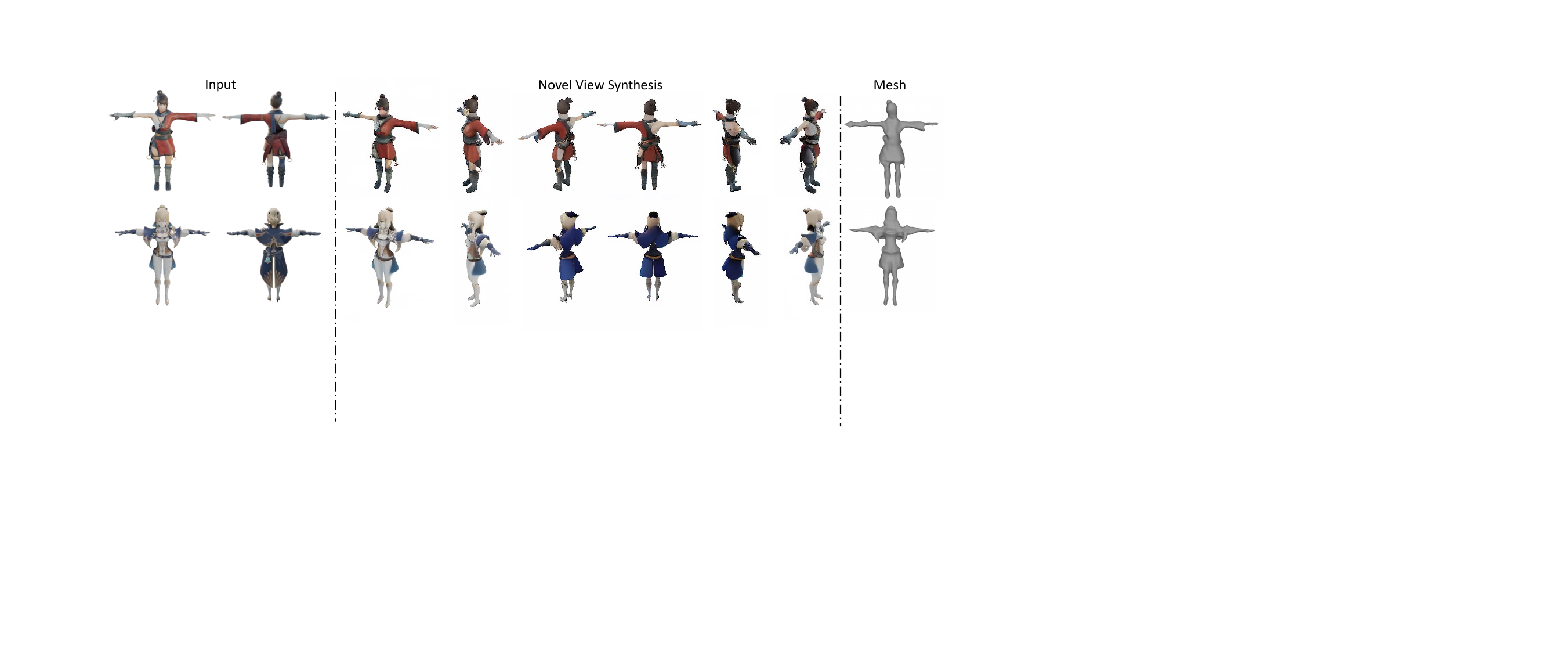}
    \caption{\textbf{3D Character Modeling.} DC-SyncDreamer is able to reconstruct arbitrary objects with rarely sparse inputs. We present the results of 3D character modeling from multi-view 2D paintings.}
    \vspace{-1.5em}
    \label{fig:control_recon}
\end{figure*}

\begin{figure}
    \centering
    \includegraphics[width=1.0\linewidth]{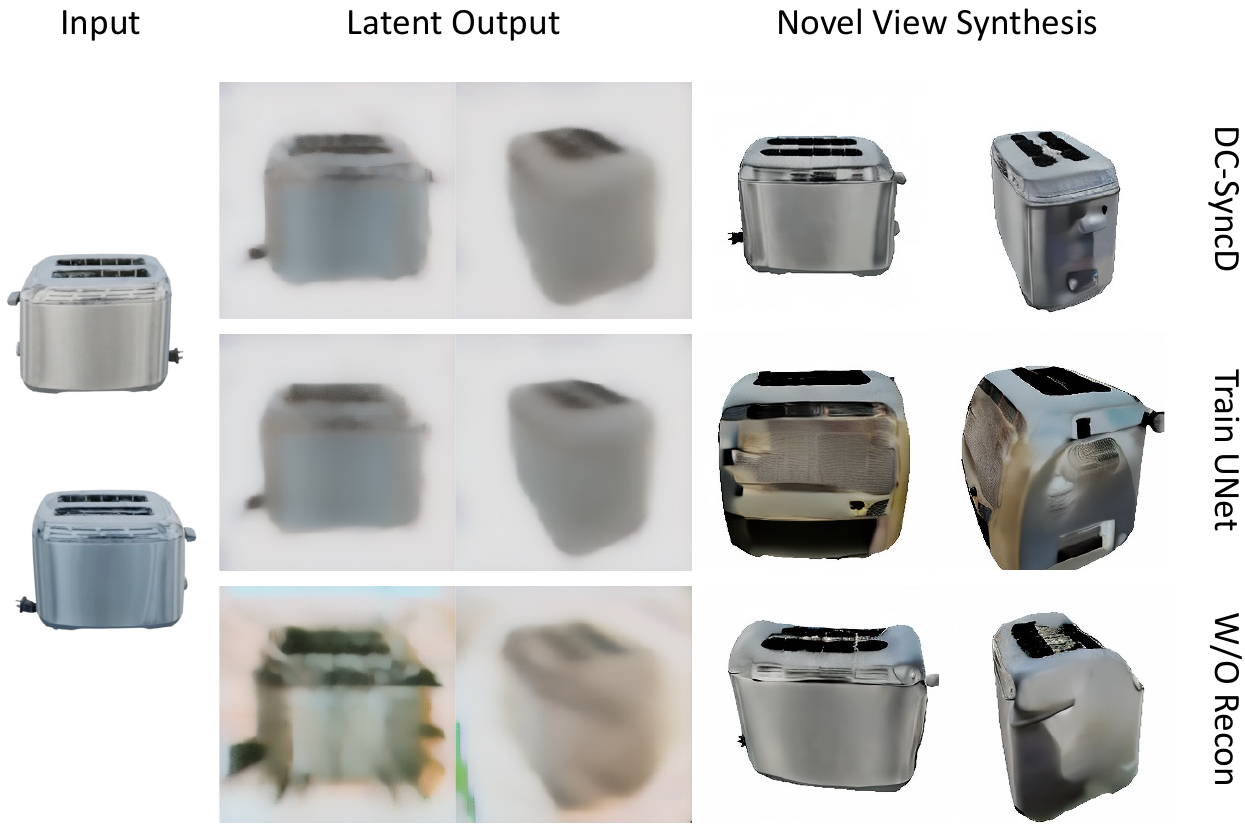}
    \caption{Ablation studies to verify the designs of our approach. ``DC-SyncD'' means our full model incorporating with SyncDreamer~\cite{liu2023syncdreamer} pipeline. ``Train UNet'' indicates finetuning the UNet with our modules without freezing it. ``W/O Recon'' means removing the reconstruction MSE loss in the second step of training. The Latent Output is derived by rendering and pooling features in the tri-planes, as shown in \figref{fig:pipeline}. 
    }
    \vspace{-1.5em}
    \label{fig:ablation}
\end{figure}

\begin{table}
    \centering
    \tablestyle{12pt}{1.08}
    \setlength{\tabcolsep}{11pt}
    \begin{tabular}{c|ccc}
       Method  & PSNR$\uparrow$ & SSIM$\uparrow$ & LPIPS$\downarrow$ \\
       \toprule
       trainable UNet
       & 15.96 & 0.762 & 0.209   \\
       w/o reconstruction loss
       & 16.18 & 0.766 & 0.206   \\
       w/o view conditioning
       & 19.04 & 0.805 & 0.166   \\
       full model 
       & \textbf{20.52} & \textbf{0.828} & \textbf{0.141}   \\
    \end{tabular}
    \caption{Ablation study on GSO dataset. Eliminating the reconstruction loss and training the UNet are both factors that negatively impact the final outcome. With view conditioning in the 3D lifting module, our model not only ensures more stable training but also yields the most optimal results.}
    \vspace{-1em}
    \label{tab:ablation}
\end{table}

\begin{figure}
    \centering
    \includegraphics[width=1.0\linewidth]{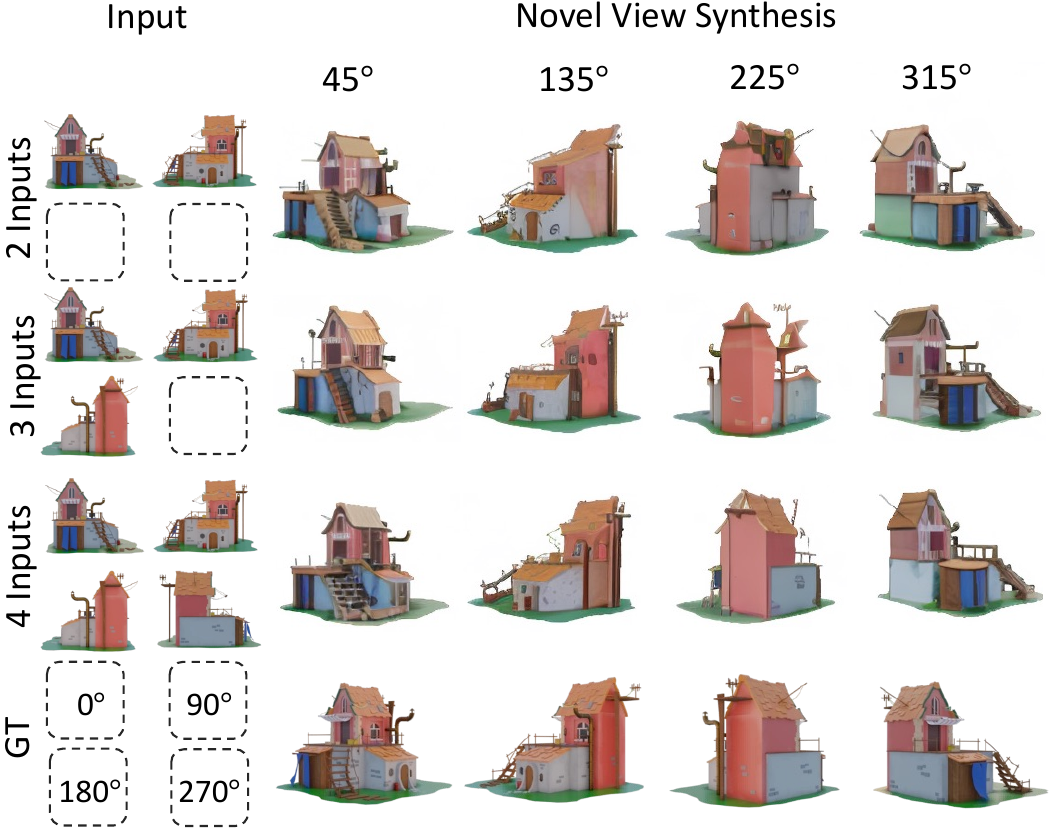}
    \caption{Ablation study to demonstrate the scalability of our model.
    Our model has the capacity to process arbitrary inputs, and its ability to control outcomes enhances correspondingly with the increasing information of input data.
    }
    \vspace{-1.5em}
    \label{fig:ablation_flex}
\end{figure}

\section{Experiments}
\label{sec:exp}
We evaluate the effectiveness of DreamComposer on zero-shot novel view synthesis and 3D object reconstruction. Datasets, evaluation metrics, and implementation details are provided in \secref{sec:datasets} and \secref{sec:imple_details}.
Our model is able to accept any number of ground-truth images as input. To show the flexibility of our framework, we integrate DreamComposer into Zero-1-to-3~\cite{liu2023zero1to3} with multi-view ground-truth inputs and SyncDreamer~\cite{liu2023syncdreamer} with the single-view ground-truth input, as described in \secref{sec:multi-view} and \secref{sec:single-view} respectively. We further demonstrate the applications of DreamComposer in \secref{sec:application}, including controllable editing and 3D character modeling. We further conduct ablation study in \secref{sec:ablation}.

\subsection{Datasets and Evaluation Metrics}\label{sec:datasets}

\textbf{Training Dataset.}
We train DreamComposer (DC) on the large-scale Objaverse~\cite{deitke2022objaverse} dataset containing around 800k 3D objects. We randomly pick two elevation angles for every object and render $N$ images with the azimuth evenly distributed in $[0^\circ,360^\circ]$. We set $N$ to $36$ for DC-Zero-1-to-3 and $16$ for DC-SyncDreamer. For training and inference, image sizes are $256 \times 256$ and background is set to white.

\noindent\textbf{Evaluation Dataset.}
To evaluate the generalization of our model to out-of-distribution data, we extend our evaluation dataset from Objaverse to Google Scanned Objects (GSO)~\cite{downs2022google}, which contains high-quality scans of everyday household items. This evaluation setting is consistent with that for SyncDreamer~\cite{liu2023syncdreamer}, comprising 30 objects that include both commonplace items and various animal species.

\noindent\textbf{Evaluation Metrics.}
Following previous works~\cite{liu2023zero1to3, liu2023syncdreamer}, We utilize Peak Signal-to-Noise Ratio (PSNR), Structural Similarity Index (SSIM)~\cite{wang2004image}, and Learned Perceptual Image Patch Similarity (LPIPS)~\cite{zhang2018unreasonable} as metrics.

\subsection{Implementation Details}\label{sec:imple_details}

During the entire training process, we randomly pick a target image as ground truth and utilize a set of three images as inputs: two images captured from opposing angles and one image from a random angle. Benefited from the image triplet training scheme, our model can adapt to two or more inputs. This data sampling strategy not only improves the efficiency of the model's optimization but also preserves its scalability and adaptability to various input configurations.

\subsection{Multi-view Input}\label{sec:multi-view}

In this section, we evaluate the performance of DreamComposer plugged into the Zero-1-to-3 with multi-view ground-truth inputs, as depicted in \figref{fig:input-num-vis} (a). 

\noindent\textbf{Evaluation Protocols.} 
When provided with an input image of an object, Zero-1-to-3~\cite{liu2023zero1to3} has the ability to generate new perspectives of the same object. 
We take the four orthogonal angles as input and set the image closest to the target perspective as the input for Zero-1-to-3 as well as the main view for DC-Zero-1-to-3. The remaining three images serve as the additional condition-views for DC-Zero-1-to-3. We calculate the results for input elevation angles of 0, 15, and 30 degrees respectively.

\noindent\textbf{Evaluation on NVS.} The comparison of quantitative results is shown in \tabref{tab:zero123_nvs}, and the comparison of qualitative results is shown in \figref{fig:zero123_nvs}. While Zero-1-to-3 possesses the ability to produce visually plausible images from novel views, the absence of multi-view inputs compromises the accuracy of these unseen viewpoints. Our DC-Zero-1-to-3, by conditioning on multi-view images, ensures the controlled generation of new viewpoints while maintaining the integrity of its diffusion model's generative capabilities. DC-Zero-1-to-3 significantly surpasses other methods in terms of the quality and consistency of generated images across various angles.

\subsection{Single-view Input}\label{sec:single-view}
In this section, we evaluate the performance of DreamComposer plugged into the SyncDreamer~\cite{liu2023syncdreamer} with single-view ground-truth inputs, as depicted in \figref{fig:input-num-vis} (b). 

\noindent\textbf{Evaluation Protocols.} We compare our method with SyncDreamer~\cite{liu2023syncdreamer}, Zero-1-to-3~\cite{liu2023zero1to3}, and RealFusion~\cite{melas2023realfusion}. Given an input image of an object, Zero-1-to-3 can synthesize novel views of the object, and SyncDreamer is able to generate consistent novel views from 16 fixed views. 
RealFusion~\cite{melas2023realfusion} is a single-view reconstruction method based on Stable Diffusion~\cite{rombach2021highresolution} and SDS~\cite{poole2022dreamfusion}. The inverse perspective of the input, generated using Zero-1-to-3~\cite{liu2023zero1to3}, serves as an additional condition-view for DC-SyncDreamer. We adhere to the identical input configurations as established in SyncDreamer. The mesh is directly reconstructed from multi-view images by NeuS~\cite{wang2023neus}. 

\noindent\textbf{Evaluation on NVS and 3D Reconstruction.} The comparison of quantitative results is shown in \tabref{tab:sync_nvs}, and the comparison of qualitative results is shown in \figref{fig:sync_nvs}. While SyncDreamer is able to generate consistent novel views, the shape of the object and the texture on the back may still appear unreasonable. DC-SyncDreamer not only maintains multi-view consistency in colors and geometry but also enhances the control over the shape and texture of the newly generated perspectives.

\subsection{Applications}\label{sec:application}
We explore the various applications of DreamComposer, including controllable 3D object editing with DC-Zero-1-to-3 and 3D character modeling with DC-SyncDreamer.

\noindent\textbf{Controllable 3D object Editing.} DreamComposer is able to perform controllable editing by modifying or designing images from certain perspectives, as shown in \figref{fig:editing}. We designate an image from a specific viewpoint as the ``control input'', which remains unaltered. Concurrently, we manipulate an ``editing input'', which represents an image from an alternate viewpoint. We utilize InstructPix2Pix~\cite{brooks2023instructpix2pix}, DragGAN~\cite{pan2023drag} and DragDiffusion~\cite{shi2023dragdiffusion} to manipulate the image, thereby achieving our desired style, corresponding to (a), (b) in \figref{fig:editing} respectively. And we modify the color of the editing input ourselves in (c). Subsequently, we employ the modified images in conjunction with the control input to synthesize novel views.

\noindent\textbf{3D Character Modeling.}
With DC-SyncDreamer, 3D characters can be modeled from only a few 2D paintings, as shown in \figref{fig:control_recon}. This can significantly improve the efficiency of existing 3D pipelines, and is expected to be connected with ControlNet for text-to-3D character creation.

\subsection{Ablation Analysis}\label{sec:ablation}
We conduct ablation studies on DC-SyncDreamer. For the ablation study, quantitative results are included in \tabref{tab:ablation} and qualitative samples are included in \figref{fig:ablation}.

\noindent\textbf{Necessity of reconstruction loss.} First, we remove the reconstruction MSE loss in the second step of training as discussed in \secref{sec:multi_view_training}. As shown in \figref{fig:ablation} and \tabref{tab:ablation}, without reconstruction MSE loss, the multi-view 3D lifting module is unable to produce effective latent outputs, resulting in the inability to synthesize satisfactory novel views.

\noindent\textbf{Finetuning v.s. freezing the diffusion U-Net.} In our design, the pretrained diffusion U-Net is frozen during training DreamController. We attempt to finetune the diffusion U-Net with DreamComposer's modules in the second stage of training. As shown in \figref{fig:ablation} and \tabref{tab:ablation}, the model performance decreases when we finetune the U-Net together with our modules. 

\noindent\textbf{Necessity of view-conditioning for 3D lifting.} We remove the view conditioning cross-attention of the 2D-to-3D lifting module. As shown in \tabref{tab:ablation}, removing the view conditioning leads to worse performance. We also empirically observe that the training is unstable without view conditioning. More results are shown in supplementary material.

\noindent\textbf{Scalability for arbitrary numbers of input views.} We validate our model's flexibility and scalability in managing arbitrary numbers of inputs. As shown in \figref{fig:ablation_flex}, our model can handle arbitrary numbers of input views, and its controllability is strengthened proportionally with the increasing number of input views.

\section{Conclusion and Discussions}
\label{sec:conclusion}

We propose DreamComposer, a flexible and scalable framework to empower existing diffusion models for zero-shot novel view synthesis with multi-view conditioning. DreamComposer is scalable to the number of input views. It can be flexibly plugged into a range of existing state-of-the-art models to empower them to generate high-fidelity novel view images with multi-view conditions, ready for controllable 3D object reconstruction. More discussions and limitations are presented in the supplementary materials. 

\section*{Acknowledgements}
\label{sec:acknow}
Thanks for the rendering data from VAST, and the discussion by Yixing Lao, Yuan Liu and Chenming Zhu.

\clearpage
\setcounter{page}{1}
\maketitlesupplementary
\appendix

\section{Implementation Details}

\begin{figure}[h]
    \centering
    \includegraphics[width=0.5\linewidth]{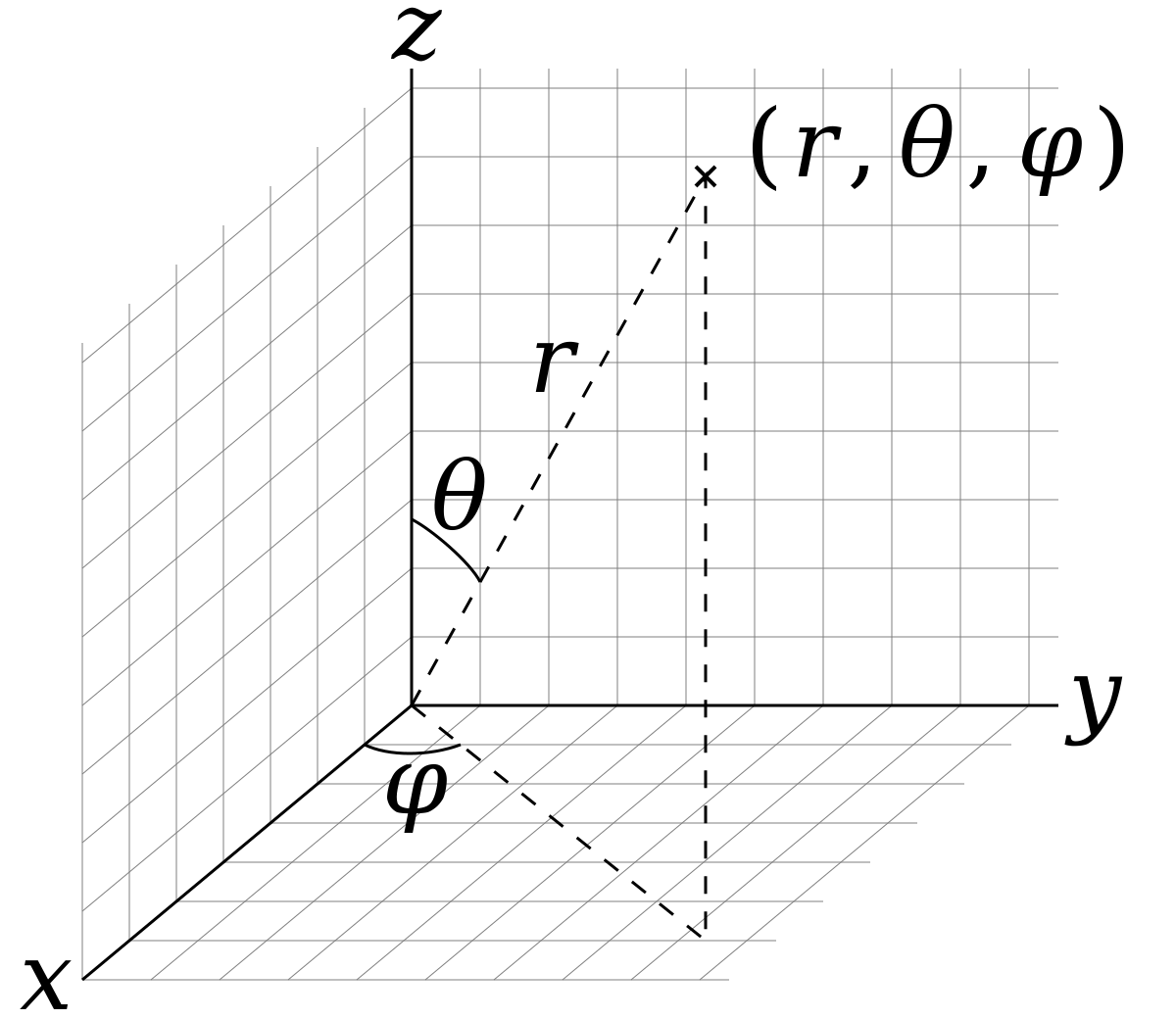}
    \caption{Spherical Coordinate System~\cite{wiki:Spherical_coordinate_system}.}
    \label{fig:cam_system}
\end{figure}

\noindent\textbf{Camera Embedding.} Following Zero-1-to-3~\cite{liu2023zero1to3}, we utilize a spherical coordinate system to represent camera locations and their relative transformations. As shown in \figref{fig:cam_system}, during the training stage, camera locations of two images from disparate viewpoints are designated as $(\theta_1, \phi_1, r_1)$ and $(\theta_2, \phi_2, r_2)$, respectively. The \emph{relative} transformation between these camera positions is expressed as $(\theta_2 - \theta_1, \phi_2 - \phi_1, r_2 - r_1)$. In both the training and inference stages, four parameters delineating the relative camera viewpoint $[\Delta\theta, \sin(\Delta\phi), \cos(\Delta\phi), \Delta r]$ are inputted into the cross-attention layers of DreamComposer's Target-Aware 3D Lifting Module and Target-View Feature Injection Module to provide camera view information.

\noindent\textbf{Architecture and Hyperparameters.} We design Target-Aware 3D Lifting Module based on the U-Net architecture from Stable Diffusion~\cite{rombach2021highresolution}. This model's architecture is specifically configured with a model dimension of 192 and includes two residual blocks at each resolution level. A distinctive feature of our approach is the integration of a cross-attention module, which facilitates the processing of relative camera embeddings.

For our experiments, we standardize the image dimensions at $256 \times 256$ pixels. Correspondingly, this establishes the latent space dimensionality at $32 \times 32$. Additionally, we configure the triplane dimensions at $32 \times 32 \times 3$, with the feature dimension of each triplane element being set to 32.

\noindent\textbf{Training Details.} We adopt a two-stage training strategy for DreamComposer. In the first stage, we focus on the 3D feature lifting module and pre-train it for 80k steps ($\sim$ 3 days) with 8 80G A800 GPUs using a total batch size of 576. The pre-trained 3D lifting module can be applied in conjunction with different pre-trained diffusion models for subsequent training. In the second stage, we jointly optimize the 3D lifting and feature injection module. This stage takes 30k steps ($\sim$ 2 days) with 8 80G A800 GPUs using a total batch size of 384.

\section{Additional Ablation Analysis}
\subsection{Comparison with NVS from sparse views}
\begin{figure*}
    \centering
    \includegraphics[width=1.0\linewidth]{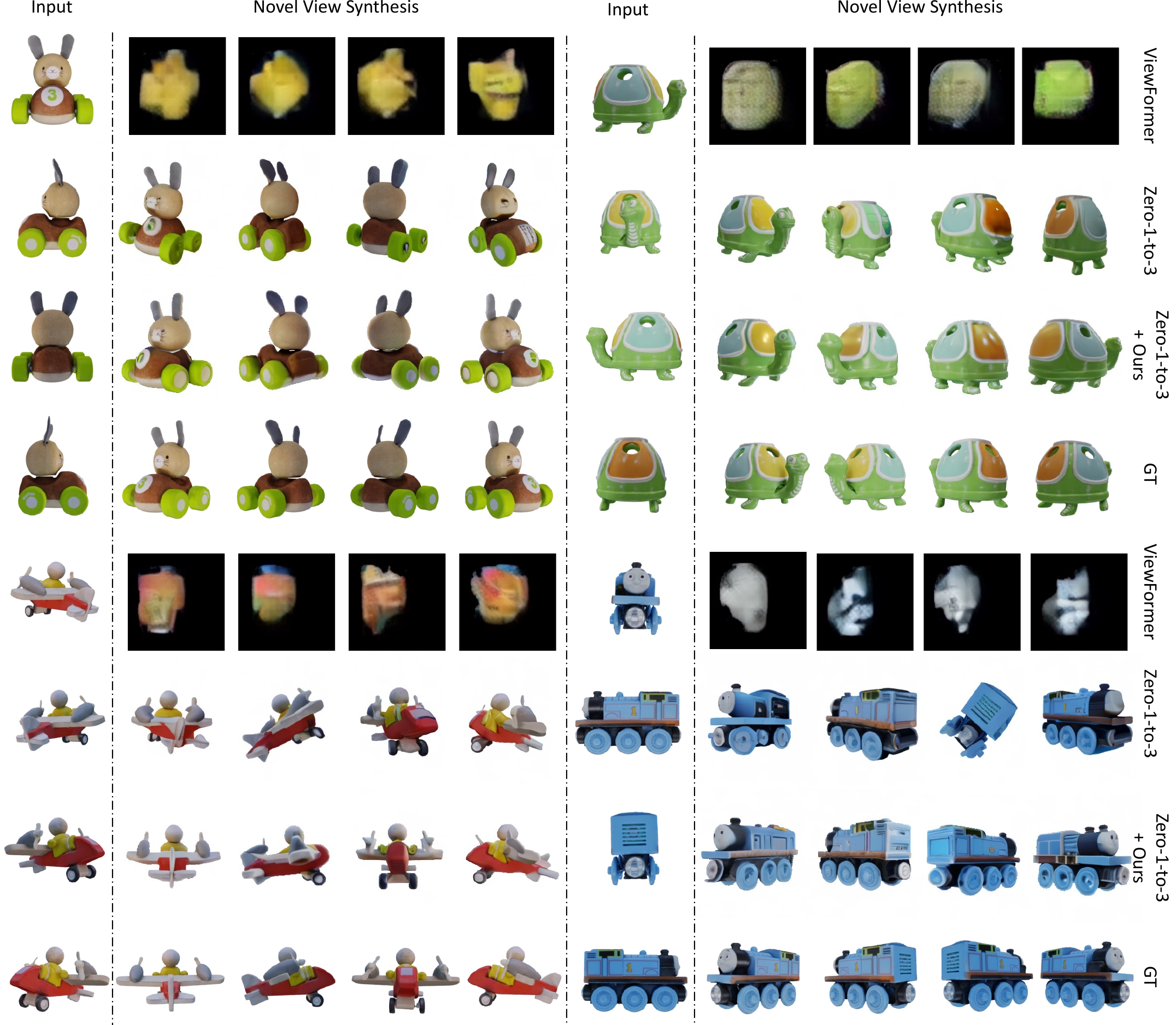}
    \caption{Qualitative comparisons with ViewFormer~\cite{kulhanek2022viewformer} and Zero-1-to-3~\cite{liu2023zero1to3} of novel view synthesis on GSO dataset using four orthogonal angles’ images as inputs. ViewFormer, despite its training on the CO3D dataset, demonstrates limitations in processing out-of-domain data. In contrast, by integrating multi-view information, our model exhibits the capability to produce controllable and superior-quality images from new perspectives of in-the-wild data.}
    \label{fig:zero123_gso_nvs_supp}
\end{figure*}

To compare with novel view synthesis methods from sparse-view inputs, we choose ViewFormer~\cite{kulhanek2022viewformer} as competitor, which achieves significant results on the CO3D dataset~\cite{reizenstein2021common}.
ViewFormer is designed for novel view synthesis using sparse-view inputs, employing transformers to process multiple context views and a query pose. This approach allows for the synthesis of novel images within an advanced neural network architecture.  For our evaluation, we utilize the ViewFormer model that has been comprehensively trained on the CO3D dataset~\cite{reizenstein2021common}, ensuring a fair comparison with its contemporary counterparts. The evaluation dataset setting is same as the one in Section 4.3. 
The quantitative results are shown in \tabref{tab:zero123_nvs_supp}, and the qualitative results are shown in \figref{fig:zero123_gso_nvs_supp}. While ViewFormer shows proficiency in handling the CO3D dataset, it exhibits limitations in processing out-of-distribution data. DC-Zero-1-to-3 far surpasses novel view synthesis methods with sparse-view inputs in both qualitative and quantitative analysis.

\begin{table}[ht]
    \begin{minipage}{0.48\textwidth}
    \subcaption{Elevation Degree - 0}\label{subtab:ed0_supp}
    \tablestyle{12pt}{1.08}
    \begin{tabular}{c|ccc}
        Methods & PSNR $\uparrow$ & SSIM $\uparrow$ & LPIPS $\downarrow$ \\
        \toprule
        ViewFormer & 13.45 & 0.630 & 0.359 \\
        Zero-1-to-3 & 20.82 & 0.840 & 0.139 \\
        DC-Zero-1-to-3 (Ours) & \textbf{25.25} & \textbf{0.888} & \textbf{0.088} \\
    \end{tabular}
    \end{minipage} \\
    \begin{minipage}{0.48\textwidth}
    \subcaption{Elevation Degree - 15}\label{subtab:ed15_supp}
    \tablestyle{12pt}{1.08}
    \begin{tabular}{c|ccc}
        Methods & PSNR $\uparrow$ & SSIM $\uparrow$ & LPIPS $\downarrow$ \\
        \toprule
        ViewFormer & 13.00 & 0.618 & 0.371 \\
        Zero-1-to-3 & 21.38 & 0.837 & 0.131 \\
        DC-Zero-1-to-3 (Ours) & \textbf{25.85} & \textbf{0.891} & \textbf{0.083} \\
    \end{tabular}
    \end{minipage} \\
    \begin{minipage}{0.48\textwidth}
    \subcaption{Elevation Degree - 30}\label{subtab:ed30_supp}
    \tablestyle{12pt}{1.08}
    \begin{tabular}{c|ccc}
        Methods & PSNR $\uparrow$ & SSIM $\uparrow$ & LPIPS $\downarrow$ \\
        \toprule
        ViewFormer & 13.02 & 0.618 & 0.373 \\
        Zero-1-to-3 & 21.66 & 0.837 & 0.128 \\
        DC-Zero-1-to-3 (Ours) & \textbf{25.63} & \textbf{0.885} & \textbf{0.086} \\
    \end{tabular}
    \end{minipage} \\
    \caption{Quantitative comparisons of novel view synthesis on 
    GSO dataset using four orthogonal angles' images as inputs. DC-Zero-1-to-3 far surpasses other methods on all metrics.}
    \label{tab:zero123_nvs_supp}
\end{table}

\begin{figure}[!h]
  \centering
  \includegraphics[width=\linewidth]{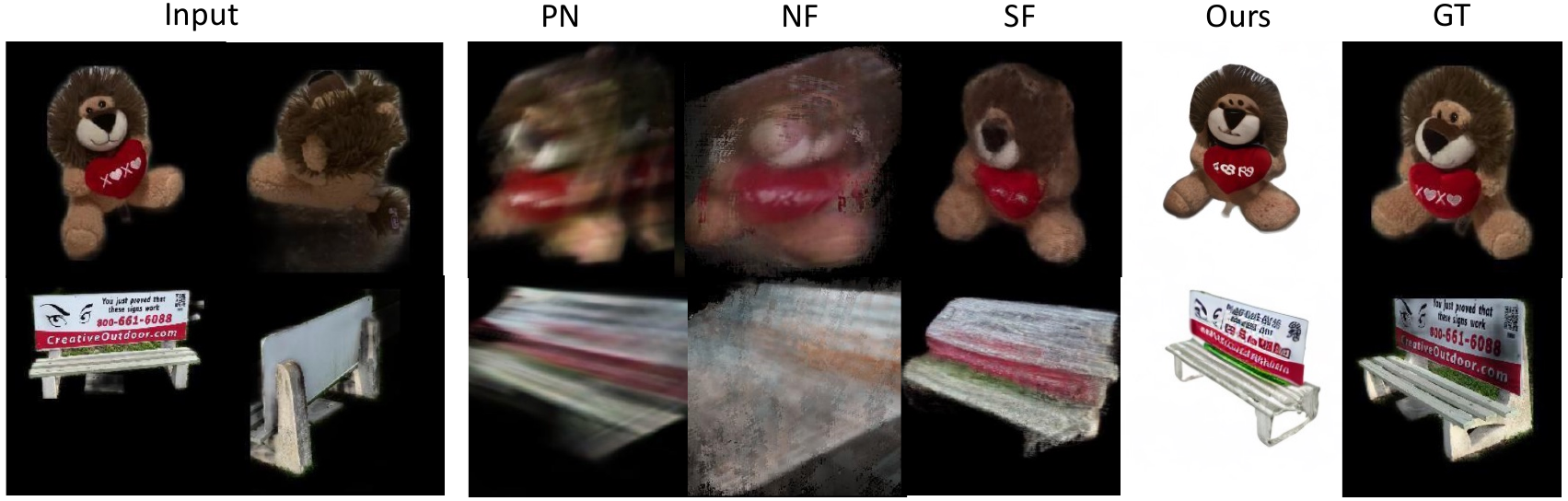}
   \caption{Qualitative comparison with PixelNeRF (PN), NerFormer (NF), SF (SparseFusion) in novel view synthesis.}
   \label{fig:co3d}
\end{figure}

Our method is capable of zero-shot learning and also demonstrates superior performance compared to other few-shot reconstruction methods when testing on their datasets. A qualitative comparison with PixelNeRF (PN), NerFormer (NF), SF (SparseFusion) is presented in \figref{fig:co3d}.
The contents in CO3DV2 dataset is not at the center of images, which is not aligned to the setting of Zero-1-to-3 and SyncDreamer, so it is not proper to give quantitative results.

\subsection{Scalability for arbitrary inputs}
\begin{figure*}
    \centering
    \includegraphics[width=0.68\linewidth]{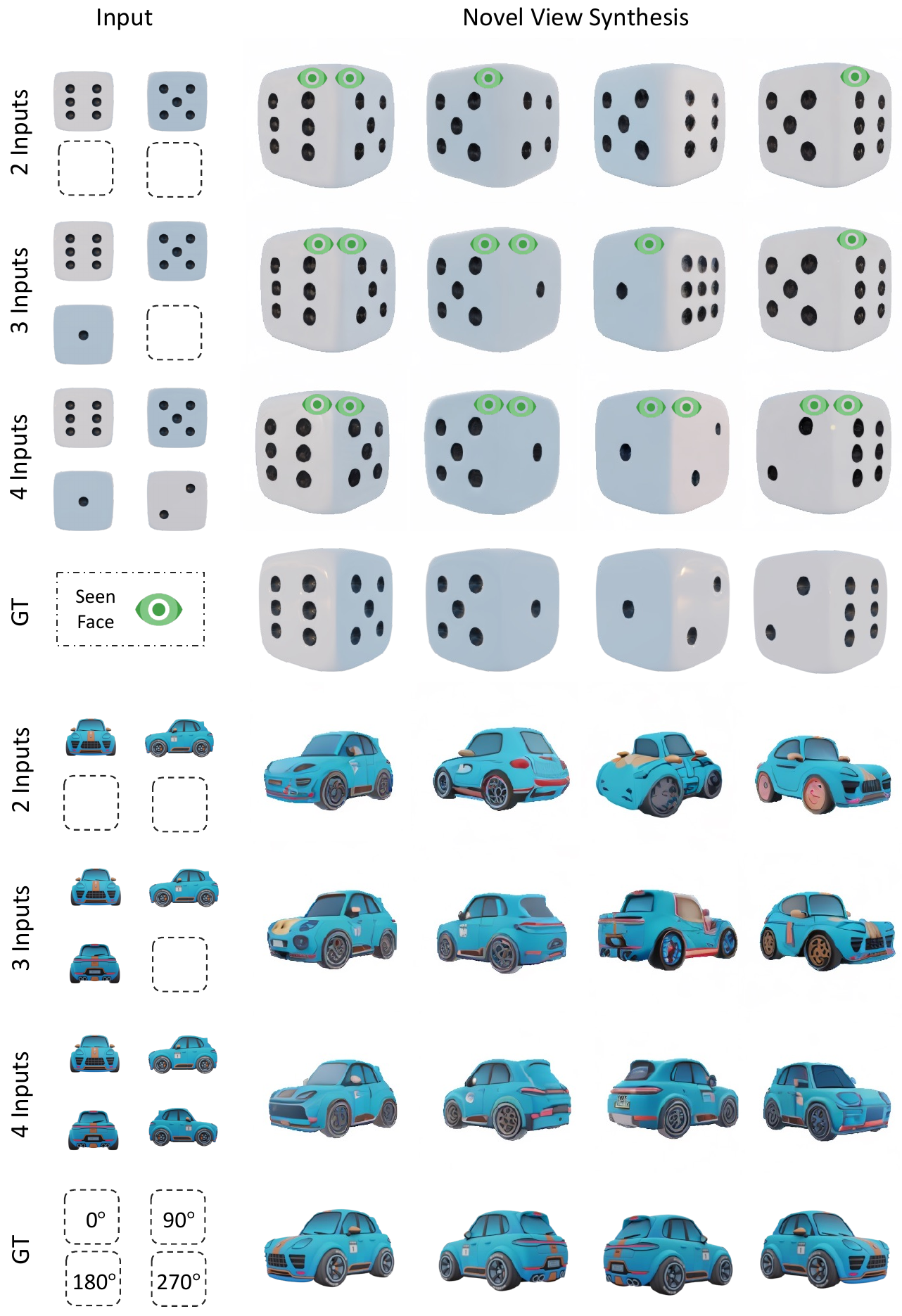}
    \caption{Ablation study to demonstrate the scalability of DreamComposer. Our model can handle various inputs and that its control over the results gets better as the amount of information from the inputs increases.}
    \label{fig:ablation_scala_supp}
\end{figure*}

We further explore our model's flexibility and scalability in managing arbitrary numbers of inputs. We evaluate the model's performance using the same set of 30 objects from Section 4.3, 
but with differing input counts. To ensure the robustness of the experiment, we strategically select input perspectives to encompass a broad area. Specifically, for 2 inputs, we use angles $0^\circ$ and $180^\circ$; for 3 inputs, we use angles $0^\circ$, $90^\circ$, $180^\circ$; for 4 inputs, we use angles $0^\circ$, $90^\circ$, $180^\circ$, $270^\circ$. We show the quantitative results in \tabref{tab:ablation_scala}. Subsequently, we assess additional datasets and present the qualitative outcomes in \figref{fig:ablation_scala_supp}.

\subsection{Necessity of view-conditioning for 3D lifting}
\begin{figure}[!h]
  \centering
  \includegraphics[width=\linewidth]{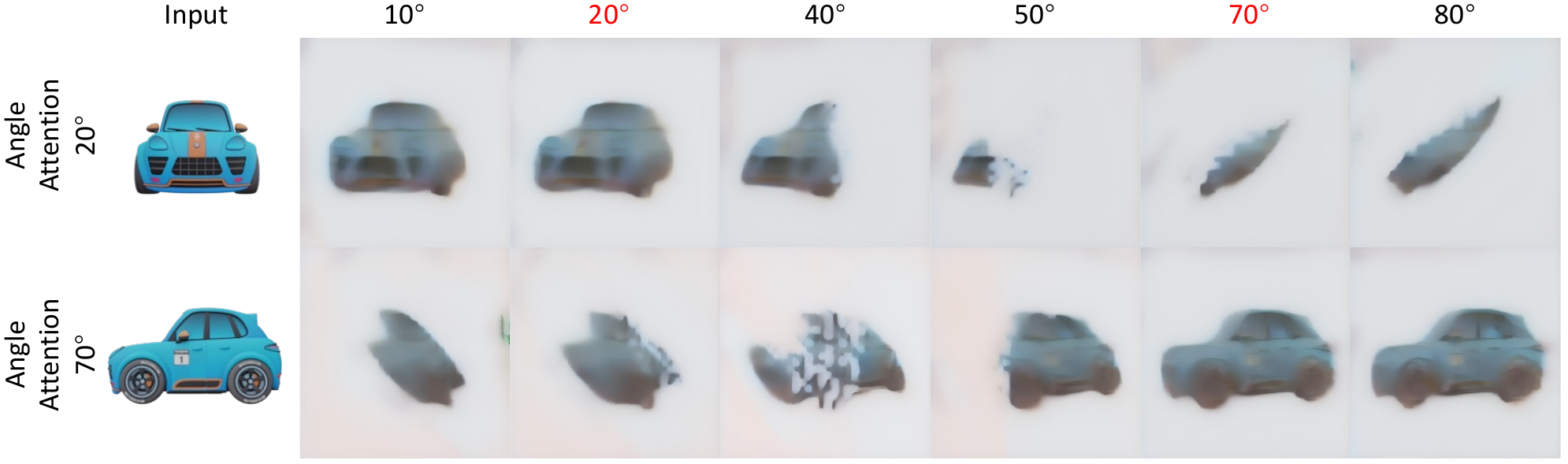}
   \caption{Latent space visualization with different angle attention.}
   \label{fig:angle_att}
\end{figure}

Under different angle difference inputs, we visualize the tri-plane features from the target view. As shown in \figref{fig:angle_att}, the projection from the target view has the highest quality. The experimental setup, as outlined in the first column, involves two inputs with the primary view presented at the top. The difference in views is computed in relation to this primary view. In the first row, the specified view difference is 20 degrees, hence, only the subsequent result at the corresponding 20 degrees is deemed valid. Similarly, in the second row, a view difference of 70 degrees is specified, making only the subsequent result at the corresponding 70 degrees valid.

\section{Additional Results}

\subsection{DC-SyncDreamer Results}
\begin{figure*}
    \centering
    \includegraphics[width=1.0\linewidth]{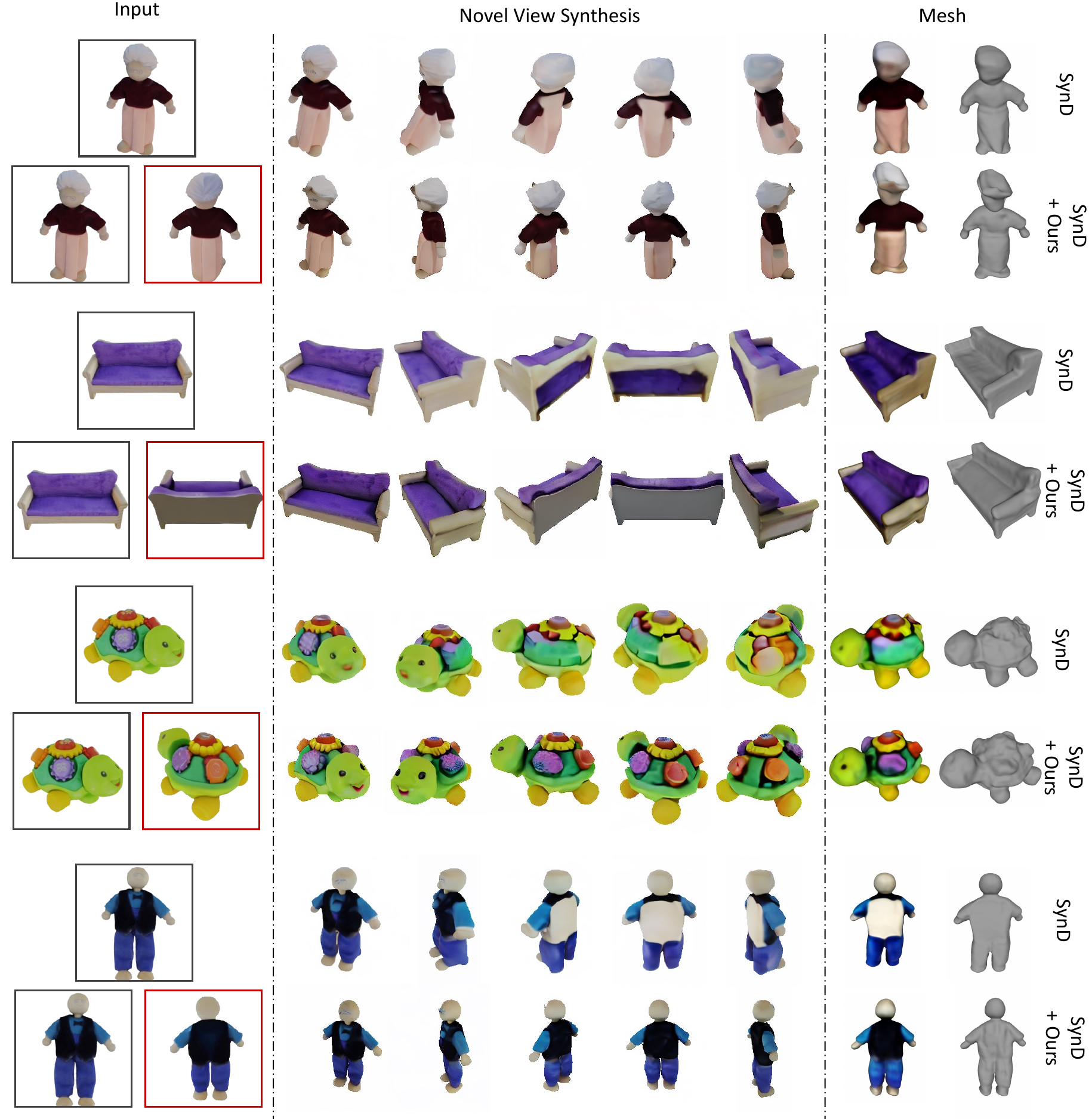}
    \caption{Qualitative comparisons with SyncDreamer (SyncD)~\cite{liu2023syncdreamer} in controllable novel view synthesis and 3D reconstruction. The image in $\square$ is the main input, and the other image in \textcolor{red}{$\square$} is the conditional input generated from Zero-1-to-3~\cite{liu2023zero1to3}. With more information in multi-view images, DC-SyncDreamer is able to generate more accurate back textures and more controllable 3D shapes.}
    \label{fig:sync_gso_nvs_supp}
\end{figure*}

\begin{figure*}
    \centering
    \includegraphics[width=1.0\linewidth]{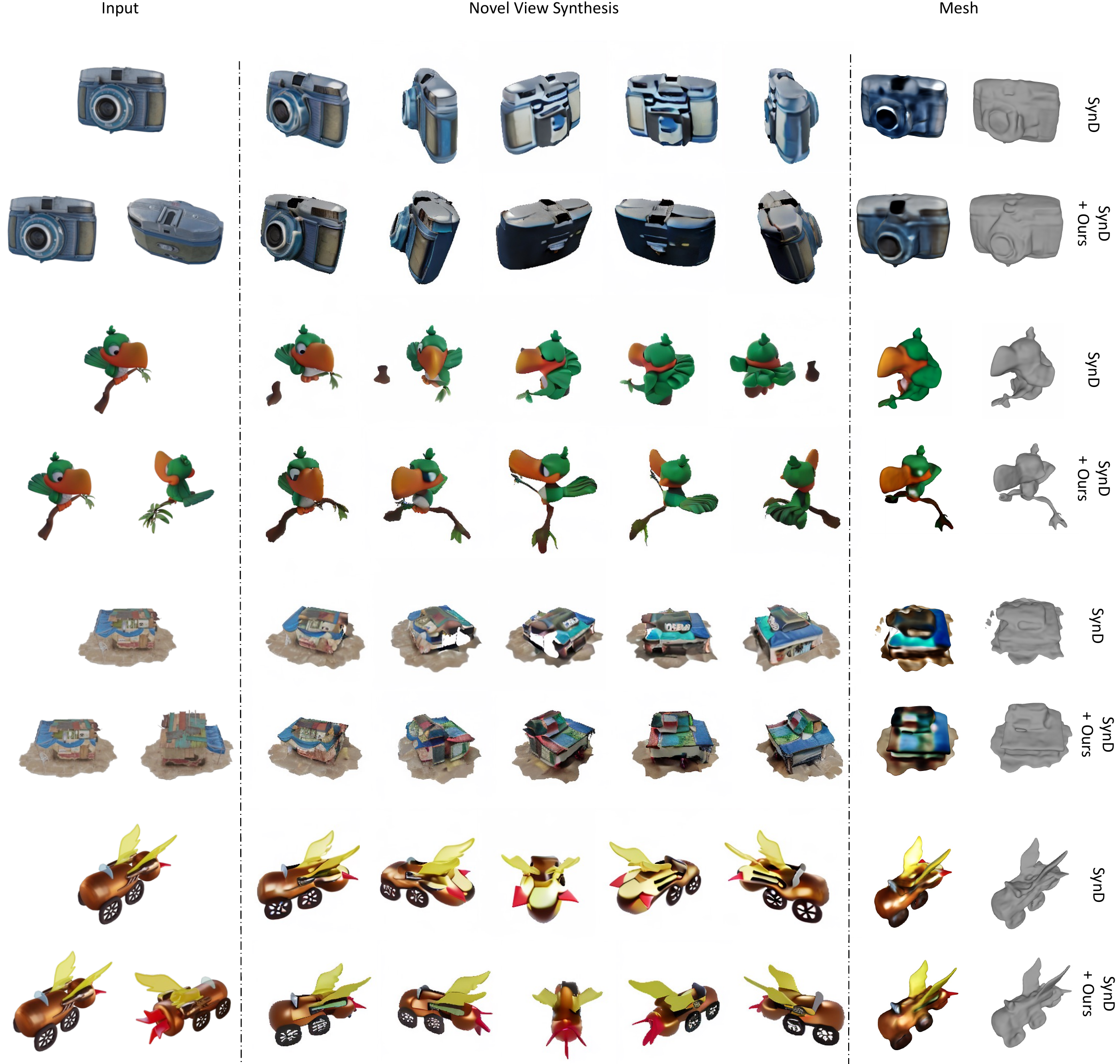}
    \caption{Qualitative comparisons with SyncDreamer (SyncD) on Objaverse dataset.}
    \label{fig:sync_obj_supp}
\end{figure*}

We present more qualitative comparisons of SyncDreamer~\cite{liu2023syncdreamer} and our method on the GSO~\cite{downs2022google} dataset, as shown in \figref{fig:sync_gso_nvs_supp}. We utilize an image as the input for SyncDreamer, as well as the main input view for our DC-SyncDreamer. We further generate the back-view image of the object with Zero-1-to-3~\cite{liu2023zero1to3}, serving as an additional condition-view for DC-SyncDreamer. 
We present additional qualitative results on Objaverse~\cite{deitke2022objaverse} dataset in \figref{fig:sync_obj_supp}. 
Leveraging the multi-view information about the object, DC-SyncDreamer is capable of generating controllable novel views and 3D objects.

\section{Limitations}
Although DreamComposer can leverage multi-view inputs to enhance zero-shot novel view synthesis, we empirically found that it is still unsatisfactory in preserving fine-grained textures from non-main view input images. It may be caused by the fact that we adopt multi-view conditioning on a low-resolution latent space, which is efficient but suffers from the loss of high-frequency details. In addition, angular deviations between multi-view input images may affect the generation quality.

\clearpage

\begin{table*}
    \centering
    \begin{tabular}{c|ccc|ccc|ccc}
       & \multicolumn{3}{c|}{Elevation Degree - 0} & \multicolumn{3}{c|}{Elevation Degree - 15} & \multicolumn{3}{c}{Elevation Degree - 30} \\
        & PSNR$\uparrow$ & SSIM$\uparrow$ & LPIPS$\downarrow$ & PSNR$\uparrow$ & SSIM$\uparrow$ & LPIPS$\downarrow$ & PSNR$\uparrow$ & SSIM$\uparrow$ & LPIPS$\downarrow$ \\
       \toprule
        2 views & 20.38 & 0.826 & 0.159 & 22.33 & 0.847 & 0.125 & 22.42 & 0.845 & 0.124 \\

        3 views & 23.68 & 0.869 & 0.108 & 24.56 & 0.875 & 0.098 & 24.27 & 0.867 & 0.102 \\
       
        4 views & 25.25 & 0.888 & 0.088 & 25.85 & 0.891 & 0.083 & 25.63 & 0.885 & 0.086 \\

        5 views & 26.10 & 0.897 & 0.081 & 26.62 & 0.899 & 0.078 & 26.52 & 0.895 & 0.079 \\
       
        6 views & \textbf{26.99} & \textbf{0.907} & \textbf{0.074} & \textbf{27.39} & \textbf{0.907} & \textbf{0.072} & \textbf{27.26} & \textbf{0.903} & \textbf{0.073}
    \end{tabular}
    \caption{Quantitative comparisons on the GSO dataset with different number of inputs. As the number of input images increases, the generation of new perspectives becomes more controllable.}
    \label{tab:ablation_scala}
\end{table*}

\clearpage

{
\small
\bibliographystyle{ieeenat_fullname}
\bibliography{main}
}

\end{document}